\documentclass[runningheads]{llncs}
\usepackage[T1]{fontenc}
\usepackage{graphicx}
\usepackage{amsmath} 
\usepackage{amssymb} 
\usepackage{booktabs}
\usepackage{hyperref}
\usepackage{multirow}
\DeclareMathOperator*{\argmin}{arg\,min}

\begin{document}
\title{POLAFFINI: Efficient feature-based polyaffine initialization for improved non-linear image registration}%\thanks{Supported by ***.}}
\titlerunning{Feature-based polyaffine initialization for non-linear registration}

\author{Antoine Legouhy\inst{1,2}, Ross Callaghan\inst{2}, Hojjat Azadbakht\inst{2}, Hui Zhang\inst{1}}
\authorrunning{A. Legouhy et al.}

\institute{Centre for Medical Image Computing \& Department of Computer Science, University College London, London, UK\and AINOSTICS ltd., Manchester, UK}

\maketitle              % typeset the header of the contribution
\begin{abstract}

This paper presents an efficient feature-based approach to initialize non-linear image registration. Today, nonlinear image registration is dominated by methods relying on intensity-based similarity measures. A good estimate of the initial transformation is essential, both for traditional iterative algorithms and for recent one-shot deep learning (DL)-based alternatives. The established approach to estimate this starting point is to perform affine registration, but this may be insufficient due to its parsimonious, global, and non-bending nature. We propose an improved initialization method that takes advantage of recent advances in DL-based segmentation techniques able to instantly estimate fine-grained regional delineations with state-of-the-art accuracies. Those segmentations are used to produce local, anatomically grounded, feature-based affine matchings using iteration-free closed-form expressions. Estimated local affine transformations are then fused, with the log-Euclidean polyaffine framework, into an overall dense diffeomorphic transformation. We show that, compared to its affine counterpart, the proposed initialization leads to significantly better alignment for both traditional and DL-based non-linear registration algorithms. The proposed approach is also more robust and significantly faster than commonly used affine registration algorithms such as FSL FLIRT.

\end{abstract}
\section{Introduction}
Medical image registration is the task of finding the best transformation, over a chosen search space, mapping a moving image onto a reference one so that the anatomical structures they portray match. In the case of non-linear registration, the search space contains non-global transformations with a high number of degrees of freedom (dof) in order to capture local, subtle displacements.

Registration techniques can broadly be divided into feature-based and intensity-based methods. 
Feature-based methods first identify common features between images, then find the transformation that best spatially aligns the corresponding features. These features are often based on geometric or anatomical characteristics which confer tangibility, interpretability. However, extracting these features historically requires tedious expert annotations or computationally intensive yet inaccurate automatic algorithms. 
On the other hand, intensity-based methods rely on a similarity measure between voxel-wise intensities of the two images as surrogate measure of quality of alignment. These approaches now dominate the field~\cite{oliveira2014}, thanks to their simple formulation as an optimization problem, the solution of which is approachable by calculus techniques like gradient descent. However, intensity similarity measures induce highly non-convex cost functions that render optimization to be prone to local minima. A good starting point is therefore essential. To this end, non-linear registration is usually preceded by an affine one and the optimization often follow a coarse-to-fine pyramidal strategy where the current step is initialized by the coarser estimate at the previous step. 
Yet, due to the global nature and limited number of degrees of freedom (dof) of affine transformations, this initialization might be insufficient.

Recently, deep-learning architectures have shown promising results in image segmentation, to the point that convolutional neural networks like U-Net now dominate challenges for this task~\cite{isensee2021}. FastSurfer~\cite{henschel2020} is able to accurately replicate FreeSurfer's segmentation - which usually takes more than 5 hours - under a minute, while SynthSeg~\cite{billot2021} is in addition able to do so in a contrast and resolution agnostic fashion. 
Anatomical feature extraction, historically difficult to achieve, is now accessible at minimal cost.

We propose, using segmentation computed from deep-learning models, to quickly produce an anatomically grounded polyaffine initial transformation as starting point for non-linear registration. The polyaffine transformation has many more dof than its affine counterpart, being able to capture non-global aspects like bends, thus offering a better start for non-linear registration algorithms. 
Based on local affine matchings with closed-form solutions, the optimization does not require iterative processes. The fusion into a dense, overall transformation is performed through the log-Euclidean polyaffine transformation (LEPT)~\cite{arsigny2009} framework which ensures a diffeomorphic result.
The end-to-end computation of the polyaffine transformation including deep-learning segmentation, local matchings and fusion onto a dense transformation can be done in less time than traditional linear registration algorithms like FSL FLIRT. 

We tested the proposed polyaffine initialization against its affine counterpart and showed that it improves the alignment of anatomical structures at end point for both traditional and deep-learning non-linear registration.  The effect was especially pronounced in the deep-learning case, where we observed a much more stable validation overlap loss during training with our approach.
%We tested the proposed polyaffine initialization against its affine counterpart using an overlap measure. Our method led to a better alignment of anatomical structures at endpoint for both traditional and deep-learning non-linear registration. It was especially pronounced in the deep-learning case where we observed a much more stable validation segmentation loss at training with our approach.
% We evaluated the relevance of our polyaffine initialization compared to its affine counterpart using an overlap measure, showing that the proposed method leads to a better alignment of anatomical structures at endpoint for both traditional and deep-learning non-linear registration. For the latter, this better starting point leads to a more stable validation segmentation loss.

\section{Method}
\label{method}
\begin{figure}
    \centering
    \includegraphics[width=\linewidth]{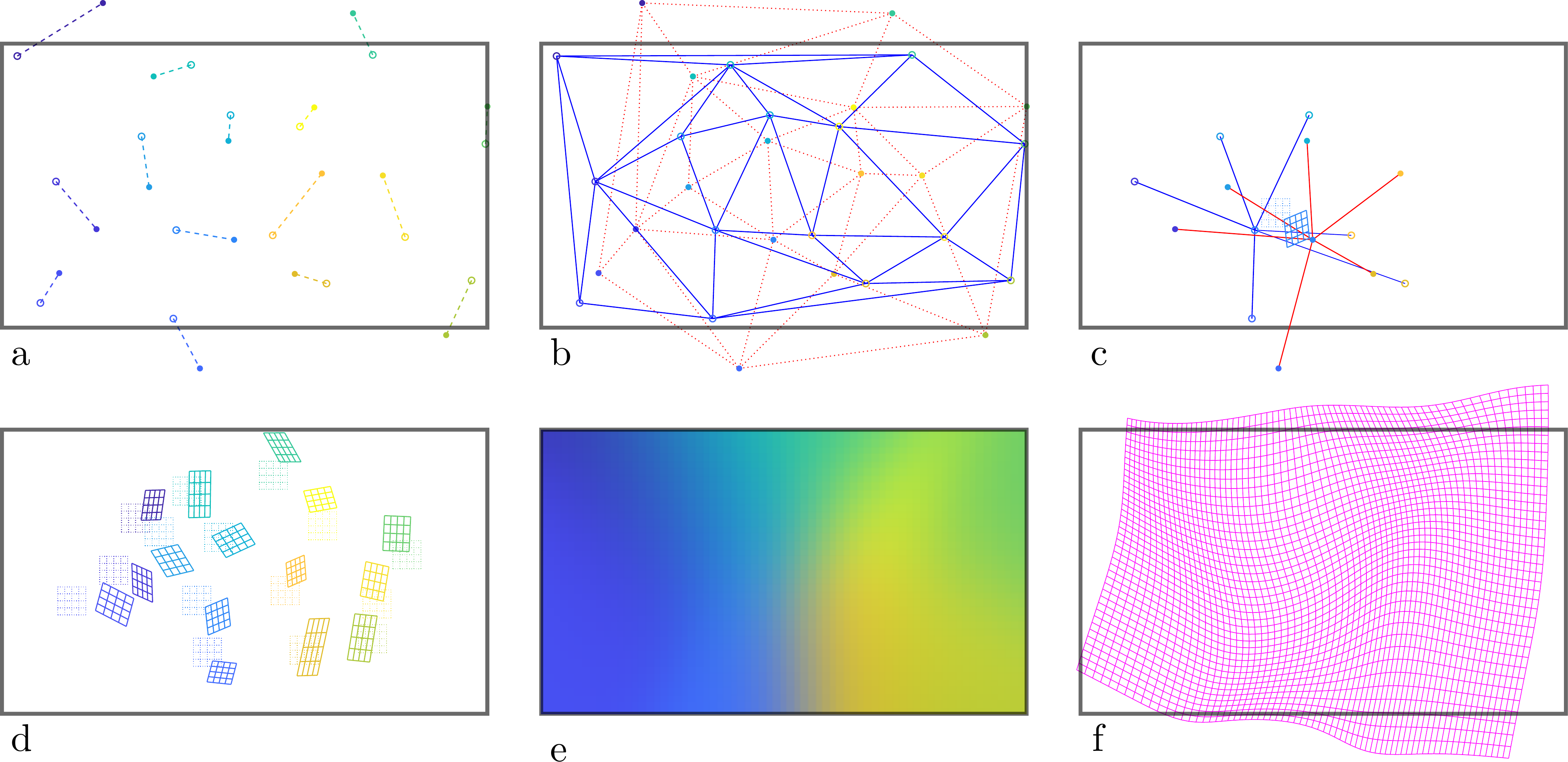}
    \caption{Illustration of several steps of the polyaffine estimation. a) Two sets of paired points (plain disc: moving, circle: reference). b) Delaunay triangulation performed on the reference set (blue) and pattern reproduced on the moving one (red). c) Local affine regression between two homologous neighborhoods. d) Set of estimated local affine transformations. e) Color grading representing all weight maps combined. f) Overall polyaffine transformation. Black rectangles represent the frame of the reference image.}
    \label{diagfig}
\end{figure}

We present here a generic method to estimate, from two sets of homologous feature points,  a dense diffeomorphic transformation under the polyaffine framework~\cite{arsigny2009}. To be anatomically grounded, the feature points are extracted from fine-grained segmentation maps, which can now be computed accurately at minimal cost thanks to freely available pre-trained deep-learning segmentation models like FastSurfer~\cite{henschel2020} or SynthSeg~\cite{billot2021}.
\begin{enumerate}
\item \textbf{Extraction of the feature points:}
Considering that the reference and moving images have undergone a fine-grained segmentation process, the feature points are defined as the centroids of the $n$ segmented regions. 
This leads to two paired sets of points (see Fig.~\ref{diagfig}.a.):
$$\left\{\begin{array}{ll}
X=\{X_1,\dots,X_n\} &\textrm{: reference point set}.\\
Y=\{Y_1,\dots,Y_n\} &\textrm{: moving point set}.
\end{array}\right.$$

\item \textbf{Estimation of the background affine transformation:} One can search for an optimal global affine transformation $\hat{A}_B$ mapping the two paired sets by formulating a linear least squares (LLS) regression problem: 
\begin{equation}
\label{llseq}
\hat{A}_B=\left(\begin{array}{cc}\hat{L}_B & \hat{t}_B\\0& 1\end{array}\right)=\argmin_{\substack{L\in \textrm{GL}_d(\mathbb{R})\\ t\in \mathbb{R}^d}}\sum_{i=1}^n\|Y_i-(LX_i+t)\|^2\end{equation}
Let us consider the relative coordinates $X'_i=X_i-\bar{X}$ and $Y'_i=Y_i-\bar{Y}$. A straightforward direct solution exists:\begin{equation}
\label{solllseq}
\hat{L}_B=\sum_{i=1}^nY'_i{X'_i}^T\left(\sum_{i=1}^nX'_i{X'_i}^T\right)^{-1} \text{\quad and\quad  } \hat{t}_B=\bar{Y}-\hat{L}_B\bar{X}
\end{equation}

Let us denote $\tilde{X}=\hat{L}_BX+\hat{t}_B$ the transformed reference feature points by the background transformation.

\item \textbf{Construction of a graph structure:} From two homologous points taken independently, one can only estimate local translations. We aim to estimate local affine transformations which each require at least 4 points in 3D (3 in 2D). To this end, we propose to also account for the contextual information from neighboring points by defining a graph structure on the reference set. 
A neighborhood, $N(X_i)$, is associated to each point $X_i$ from the reference set which is a collection of neighbors from the same set, chosen according to some suitable criterion, e.g. spatial distance or region adjacency. Since the moving and reference sets are paired, the graph structure only has to be defined on one of them. To simplify notations, we define $N(i)$ as the set of neighbor indices of $X_i$ such that: $p\in N(i)\equiv X_p\in N(X_i)$. In our (3D) implementation, we chose to define the graph structure using a Delaunay triangulation (see Fig.~\ref{diagfig}.b) which is very easy and fast to compute.
In this construction, two elements $X_i$ and $X_j$ are neighbors if they are connected by an edge. Since each $X_i$ is the vertex of one or more tetrahedrons with other elements of $X$, there is a minimum of 4 points per neighborhood, enough to regress an affine transformation. 
 
\item \textbf{Estimation of the local affine transformations:} For each $\tilde{X}_i$, a local optimal affine transformation $\hat{A}_i$ matching its neighborhood to the homologous neighborhood of $Y_i$ is sought through LLS in the same vein as in Eq.~\ref{llseq} (see Fig.~\ref{diagfig}.c): 
\begin{equation}
\label{affreg}
\hat{A}_i=\left(\begin{array}{cc}\hat{L}_i & \hat{t}_i\\0& 1\end{array}\right)=\argmin_{\substack{L\in \textrm{GL}_d(\mathbb{R})\\ t\in \mathbb{R}^d}} \sum_{p\in N(i)}\|Y_p-(L\tilde{X}_p+t)\|^2
\end{equation}
The solution is similar to Eq.~\ref{solllseq}, except that the sum and averages are done using only the points of a neighborhood rather than using all points. 
At this stage we have $n$ local affine transformations between homologous neighborhoods attached to the $n$ reference feature points (see Fig.~\ref{diagfig}.d).

% Given the homologous neighborhood centroids $\bar{X}_i=\frac{1}{|N(i)|}\sum_{p\in N(i)}X_p$ and $\bar{Y}_i=\frac{1}{|N(i)|}\sum_{p\in N(i)}Y_p$, let consider the new coordinates $X_p^i=X_p-\bar{X}_i$ and $Y_p^i=Y_p-\bar{Y}_i$. A straightforward direct solution exists:
% \begin{equation}
% \hat{L}_i=\sum_{p\in N(i)}Y_p^i{X_p^i}^T\left(\sum_{p\in N(i)}X_p^i{X_p^i}^T\right)^{-1} \text{ and } \hat{t_i}=\bar{Y}_i-\hat{L}_i\bar{X}_i
% \end{equation}

\item \textbf{Creation of the weight maps:}
To create an overall dense transformation, weight maps are established to spatially modulate the contribution of each local affine transformations. For any point $x$ of the domain over which the image is defined, a set of weights associated to each $X_i$ is defined using a smooth kernel function based on the distance between $x$ and the center of the associated neighborhood $\bar{X}_i$, and a parameter $\sigma$ controlling the smoothness (see Fig.~\ref{diagfig}.e). The kernel can typically be Gaussian of standard deviation $\sigma$.
% \begin{equation}w_i(x)=K\left(\frac{\left\|x-\bar{X}_i\right\|}{\sigma}\right)\end{equation}
The set $\{w_i(x),\enskip i=1,\dots,n\}$ can be sparse if the kernel has a bounded support. In addition, a background weight $w_B$, uniform across the whole image domain, is chosen. It should be small enough to be negligible with respect to the weights associated to the feature points.

% \begin{equation}
% w_i(x)=\frac{1}{\sqrt{2\pi}\sigma }\exp\left(-\frac{1}{2}\frac{(x-\bar{X}_i)^2}{\sigma^2}\right) \text{\quad and \quad} w_B(x)=w_B
% \end{equation}

\item \textbf{Construction of the overall dense diffeomorphic transformation:} From the collection of local affine transformations and the weight maps, one can produce an overall dense diffeomorphic  transformation through the log-Euclidean polyaffine framework~\cite{arsigny2009}. 
\begin{enumerate}
    \item A stationary velocity field (SVF) $V$ is built by interpolating a log displacement vector for each $x$ by averaging the logarithms of the local transformations, weighted by the associated weight maps and the background weight:
    \begin{equation}\label{logmean}V(x)=\frac{\sum_{i=1}^n w_i(x).\log(\hat{A}_i)}{w_B+\sum_{i=1}^nw_i(x)}x\end{equation}

    \item A diffeomorphic transformation $\exp(V)$ can be obtained by integration of a stationary ordinary differential equation (ODE) (see Fig.~\ref{diagfig}.f):
    \begin{equation}
    \label{ode}
    \exp(V)=\phi^1
    \text{, where } \left\{
    \begin{array}{l}\frac{\partial \phi^t}{\partial t}=V(\phi^t)\\ \phi^0=\mathrm{Id}\end{array}\right.
    \end{equation}
    The integration can be done efficiently on regular grids using the scaling and squaring method~\cite{arsigny2009} by approximating the exponential of the scaled (close to 0) field and composing recursively.
    \item This diffeomorphic transformation composed after the background affine one forms the overall transformation: $
    T=\hat{A}_B\circ \exp(V)$.
    
    \end{enumerate}
\end{enumerate}

\begin{remark}
For more robustness, one can imagine weighted or trimmed versions of the linear least squares minimization in Eq.~\ref{affreg} if the graph defined at step~2 is weighted, or to account for uncertainty in the feature point extraction.
\end{remark}

\begin{remark}
When dealing with smaller neighborhoods, one can regress local transformations with less degrees of freedom. For a rigid transformation, i.e. $L$ constrained to be a rotation matrix, a minimum of 2 points in each neighborhood are needed (in 3D), and a direct solution to the LLS problem can be found in~\cite{horn1987,horn1988}. For translations only, i.e. $L$ an identity matrix, the solution is simply $\hat{t}=\bar{Y}-\bar{X}$. Only singleton neighborhoods are necessary, i.e. $N(X_i)=X_i$. However, this setting reduces robustness that comes with having more equations than unknowns, and the contextual information from the neighboring positions is discarded (although it still has an impact during interpolation).
\end{remark}
\begin{remark}
\label{rmkweightmap}
The choice of the weight maps is crucial to shape the overall transformation. The window of the kernel modulates the amount of smoothness. For small values of $\sigma$, one can recover very local changes but there is a risk of overfitting or having sub machine precision values when the feature point set is too sparse in space (although the use of a background weight helps). As $\sigma$ increases, the polyaffine result gets smoother, losing curvature, eventually converging to an affine transformation for $\sigma=\infty$. This spectrum of possible transformations is a strength of the method.   
For Gaussian kernels, we found empirically that taking $\sigma=\frac{2}{n}\sum_{i=1}^n\argmin_{X_p\in X}\|X_i-X_p\|$ often leads to good results.
\end{remark}
\begin{remark}
The role of the background transformation is to ensure stability when extrapolating in regions far away from the feature points. As this polyaffine mapping doesn't assume any pre-alignement, it cannot converge to null displacement towards the domain boundary. Instead, it should follow the overall flow in the smoothest way i.e. through a global affine. 
%An background transformation defined as: $\hat{A}_B=\frac{1}{n}\exp\left(\sum_{i=1}^n\log(\hat{A}_i)\right)$ would also perfectly make sense. 
The background weight should be sufficiently small to only matter far enough from the feature points but not interfere near them. 
\end{remark}
\begin{remark}
In Eq.~\ref{logmean}, we used a log-Euclidean average of the affine transformations, i.e. an Euclidean mean on their principal logarithms, to ensure an invertible overall transformation. This invertiblity wouldn't be guaranteed with a simple Euclidean average. The well-definiteness of the principal logarithm of an affine transformation matrix $A$ only depends on its linear part $L$. The eigenvalues of $L$ must not lie on the (closed) half-line of negative real numbers~\cite{cheng2001}. This, however, only concerns very large transformations that effect a rotation close to $\pi$, which is unlikely in real settings.
There are only $n$ matrix logarithms computation to perform which can be done efficiently through the inverse scaling and squaring method for matrices~\cite{higham2008}. 
\end{remark}

\section{Experimental design}
To evaluate the benefit of the proposed segmentation-based polyaffine initialization compared to its affine counterpart for traditional and deep-leaning non-linear registration, we registered subjects from 3 databases onto a template and computed a structure overlap measure. 
\subsection{Data}
We used T1-weighted images from 3 databases in order to cover various acquisition protocols and to have brains of different maturation and health conditions:
\begin{itemize}
\item ADNI (\url{adni.loni.usc.edu}), the Alzheimer's Disease Neuroimaging Initiative~\cite{adni}, is a cohort of elderly subjects divided into cognitively normal (HC), with mild cognitive impairment (MCI), and with Alzheimer's disease (AD). 
\item IXI dataset (\url{brain-development.org/ixi-dataset}) is composed of adult healthy subjects aged 20 years old or more.
\item UK Biobank (\url{ukbiobank.ac.uk}) is a huge database of subjects from the UK, between 40 and 69 years old.
\end{itemize}
For all databases, the voxel size is around 1 mm isotropic. The reference for registration is an MNI template (ICBM 2009a~\cite{fonov2011}) with 1 mm isotropic voxel size.
\begin{table}[]
\centering
\begin{tabular}{c||c|c|c}
Database & Training set & Validation set & Testing set\\\hline\hline
IXI   & 20  & 5  & 100 \\
UK Biobank & 20   & 5  & 100   \\
\begin{tabular}{c}ADNI\vspace{-0.1cm}\\ \scriptsize{(HC/MCI/AD)}\end{tabular} & \begin{tabular}{c}60\vspace{-0.1cm}\\ \scriptsize{(20/20/20)}\end{tabular}  & \begin{tabular}{c}15\vspace{-0.1cm}\\\scriptsize{(5/5/5)}\end{tabular} & \begin{tabular}{c}150\vspace{-0.1cm}\\\scriptsize{(50/50/50)}\end{tabular}
\end{tabular}
\label{datasplit}
\caption{Distribution of the subjects for the training, validation and testing sets.}
\end{table}
Subjects have been drawn randomly from those databases and distributed into training, validation and testing sets following  Table~\ref{datasplit}. 
\subsection{Segmentation}
We used FastSurfer~\cite{henschel2020} to quickly (in less than a minute) produce FreeSurfer-like segmentations of the moving and reference images into 95 anatomical regions following the Desikan–Killiany–Tourville (DKT) \cite{klein2012,desikan2006} protocol.

\subsection{Registration}
\subsubsection{Affine initialization:} The affine pre-registrations were performed using FSL FLIRT~\cite{flirt}. For 37 subjects in all sets combined ($7.8\%$), FLIRT optimization failed, leading to severe misalignments like upside-down brains. To obtain decent outputs, we re-ran those that failed with a little "help" like resampling onto the reference image grid or constraining the search space. 

\subsubsection{Polyaffine initialization:}
From the general recipe presented in Section~\ref{method}, we opted for the following implementation details:
\begin{enumerate}
\item \textbf{Extraction of the feature points:}
Four regions were ignored: left and right cerebral white matter (too large), white matter hypointensities and cerebrospinal fluid (not consistent between images). For each remaining region, a centroid (feature point) was computed as the average spatial coordinates of the voxels belonging to the region. 

\item \textbf{Construction of the graph structure:}
The Delaunay triangulation was computed only once on the template feature points using the Qhull library.

\item \textbf{Estimation of the local affine transformations:}
Optimal affine transformations between homologous neighborhoods were computed using the direct solution in Eq.~\ref{solllseq}. 
\item \textbf{Creation of the weight maps:}
We opted for a Gaussian kernel of standard deviation $\sigma=20$ mm roughly following the rule of thumb in Remark~\ref{rmkweightmap}. We set the background weight to $w_B=10^{-5}$ uniformly across the image domain. % which is equivalent in that case to the weight at around 8 cm from a control point. 
\item \textbf{Construction of the overall dense diffeomorphic transformation:}
We computed the SVF on the image grid downsampled by a factor 2 to quicken the subsequent exponentiation.
The exponential of the SVF in Eq.~\ref{ode} was computed through scaling and squaring using 7 integration steps and resampled onto the original grid.
\end{enumerate}
All polyaffine initializations worked fine at first attempt. We have made freely available on Github\footnote{\url{https://github.com/CIG-UCL/polaffini}} the implementation described here, which is based on the Python version of SimpleITK wrapper for ITK open-source software. 

\subsubsection{Traditional non-linear registration:}
Symmetric Normalization (SyN)~\cite{ants}, from the Advanced Normalization Tools (ANTs) suite was used. It is one of the best traditional non-linear algorithms according to the evaluation in~\cite{klein2009}. The optimized similarity metric was the local squared correlation coefficient (LCC).

\subsubsection{Deep-learning non-linear registration:} Voxelmorph~\cite{balak2019} style architectures with diffeomorphic implementations~\cite{dalca2019} were used. 
They are composed of a U-Net shaped as in~\cite{balak2019}, containing all the trainable parameters, that takes as input a moving and a reference image, and outputs a vector field. This vector field is fed to an integrator block to produce a diffeomorphism which is used to transform the moving image through a resampler block.  We chose LCC as image similarity loss with weight 1. We used segmentations as auxiliary data to compute an average Dice overlap score used as segmentation loss with weight 0.3. A regularization loss, L$^2$ norm of the Jacobian of the vector field was also used to promote smooth estimates, with weight 1. We trained one model using images pre-registered using the proposed polyaffine initialization, and a second one using its affine counterpart. Due to limited GPU memory (8 GB), the models operated on images downsampled to $2\times 2\times 2$ mm grids. 

\subsection{Evaluation}
The evaluation was performed on the subjects from the testing set that were unseen by the deep-learning models at training. 
For each image, the transformation estimated at initialization and the one from the non-linear registration were applied together at once so that the transformed moving image was reconstructed in the reference grid with a single interpolation.
The quality of registrations was evaluated using a Dice overlap score between the segmentation labels of the reference image and the corresponding ones of the transformed moving image.
For the sub-cortical results, we computed 1 dice per region and reported the average. For the cortex, in order to be comparable with~\cite{balak2019}, we regrouped the regions into one label and computed a single Dice.

Having a good overlap score between the reference and the transformed moving segmentation is meaningless if it is the result of a non topology-preserving process. Under reasonable conditions (rotations with magnitude smaller than $\pi$), polyaffine transformations under the LEPT framework~\cite{arsigny2009} are diffeomorphisms. Also, the examined non-linear registration algorithms (ANTs and deep) contain a regularization term encouraging smooth transformations, yet not completely forbidding them to be improper. To evaluate the topology-preserving aspect, we also  counted the number of negative values of the Jacobian determinant of each estimated final transformation (initialization transformation composed with non-linear registration output).

\section{Results}
\begin{table}
\renewcommand{\arraystretch}{1.1}
\setlength{\tabcolsep}{5pt}
\centering
\begin{tabular}{c||l|l|l|c}
& affine init. & polyaffine init. & p-value & Cohen's d\\\hline\hline
\multirow{3}{*}{\begin{tabular}{c}sub-cortical\\ regions\end{tabular}} & affine &  polyaffine &  $< 10^{-115}$ & 1.864 \\
& affine + ANTs &  polyaffine + ANTs &  $< 10^{-14}$ & 0.447  \\
& affine + deep &  polyaffine + deep &  $< 10^{-108}$ & 1.762 \\\hline
\multirow{3}{*}{cortex} & affine &  polyaffine & $< 10^{-107}$ & 1.745 \\
& affine + ANTs &  polyaffine + ANTs & $< 10^{-38}$ & 0.798 \\
& affine + deep &  polyaffine + deep & $< 10^{-176}$ &  3.019 \\
\end{tabular}
\label{tabstat}
\caption{Statistics about differences between overlap scores after registration with affine vs proposed polyaffine initialization. Reported p-values are for paired t-tests, Cohen's d for paired samples (with polyaffine $-$ affine on numerator).}
\end{table}

\begin{figure}
    \centering
    \includegraphics[width=\linewidth]{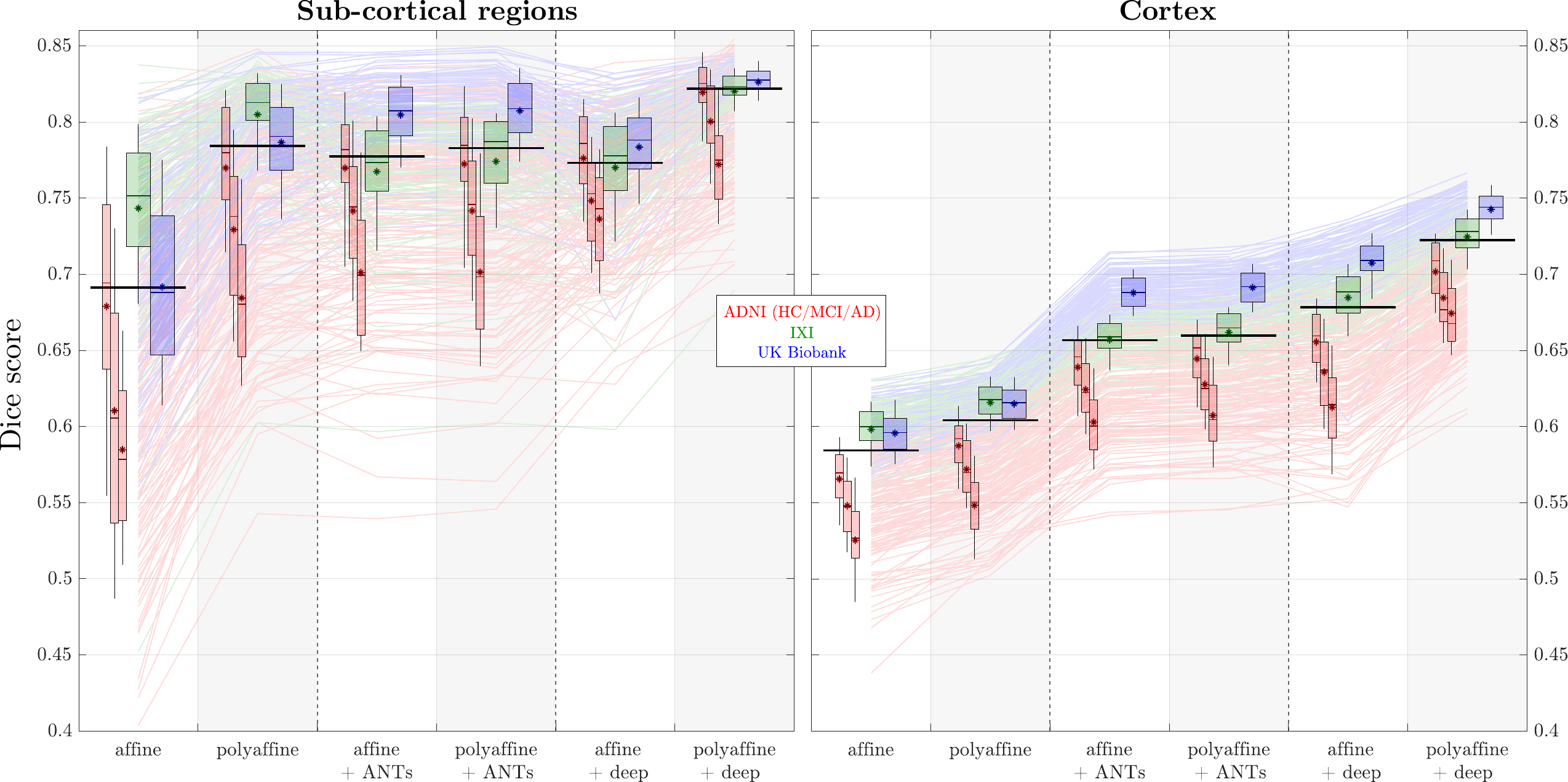}
    \caption{Dice scores after non-linear registration initialized with affine and proposed polyaffine transformation for ADNI (red, HC/MCI/AD left to right), IXI (green) and UK Biobank (blue) subjects. Thick lines represent medians across all datasets.}
    \label{dicefig}
\end{figure}

Dice scores for sub-cortical areas and cortex for the various datasets are shown in Fig.~\ref{dicefig}. Statistics regarding differences between overlap scores after registration with affine against proposed polyaffine initialization are reported in Table~\ref{tabstat} for all datasets combined. 

We observe that the sub-cortical structures are generally well aligned already just with the polyaffine initialization. While non-linear registrations do not typically improve upon this initial alignment, the deep learning model with polyaffine initialization shows clearly better overlaps.
For a given approach (ANTs or deep), results are significantly better when initializing with the proposed approach, with a large effect size for the deep-learning models. 

In the cortex, non-linear registration clearly improves upon the initial alignment. Using the proposed polyaffine initialization once again leads to significantly better results. While the effect appears modest for ANTs, it is consistent over almost all subjects. The difference is however striking for the deep-learning models where we report a very large effect size.

\begin{figure}
    \centering
    \includegraphics[width=\linewidth]{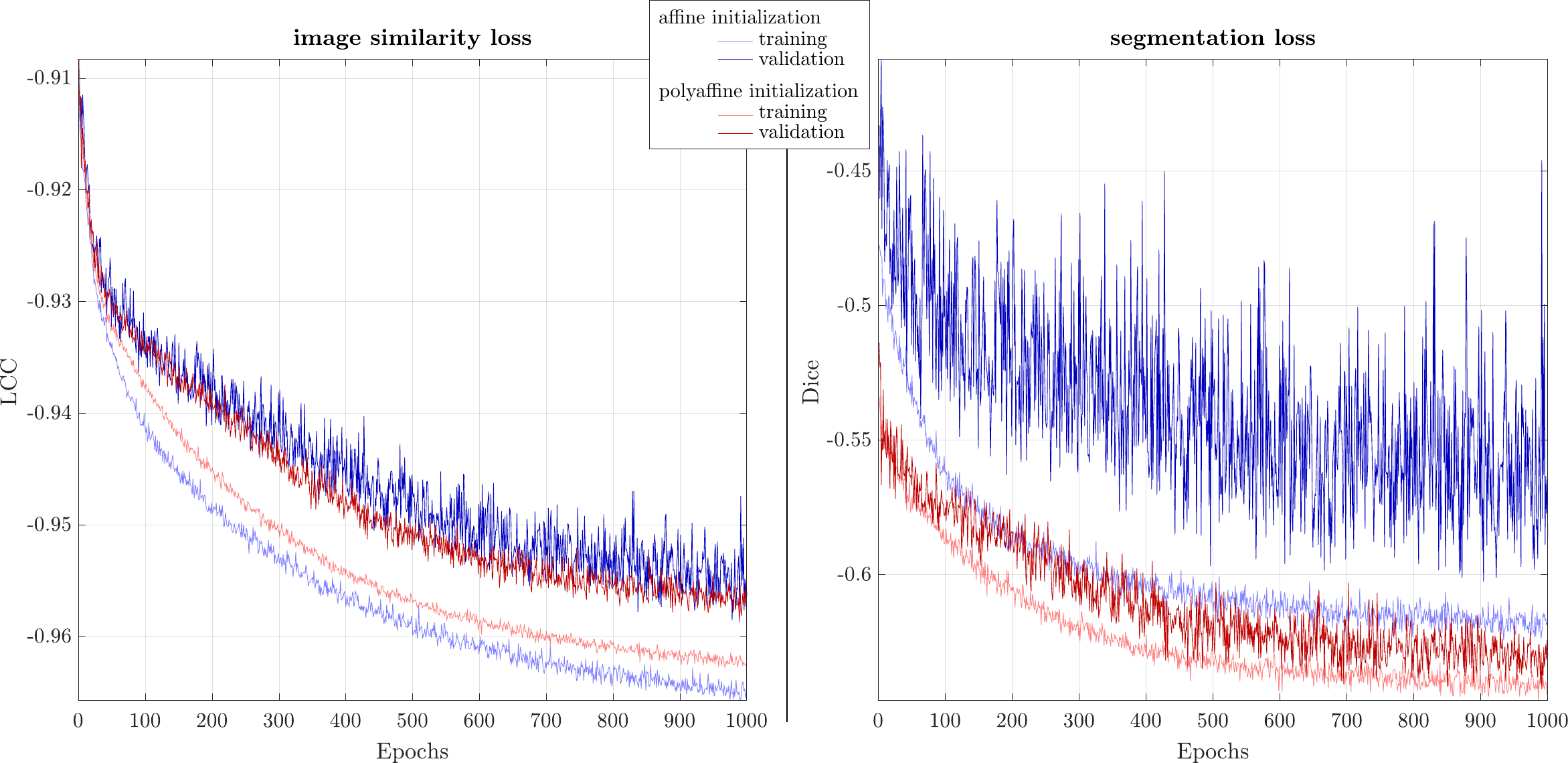}
    \caption{Evolution of image similarity (LCC) and segmentation (average Dice on all regions) losses during training of deep-learning models with affine (blue) and proposed polyaffine (red) initialization for training (light) and validation (dark) samples.}
    \label{lossfig}
\end{figure}

Part of the explanation for why the proposed polyaffine initialization leads to better results in the deep-learning case can be found by examining the evolution of the losses during training~\ref{lossfig}.
While image similarity losses follow similar trajectories for both approaches, segmentation losses, which actually quantifies the alignment of anatomical structures, show a much smoother profile with the proposed polyaffine starting point. The gap between the training and validation losses is also much smaller with the proposed initialisation.

The resulting overall transformations were proper for all subjects and methods. We did not find any negative Jacobian determinant whether using an affine or the proposed polyaffine initialization and whether using ANTs or a deep-learning model for the subsequent non-linear registration. 

% \begin{figure}
%     \centering
%     \includegraphics[width=\linewidth]{result_img.pdf}
%     \caption{Transformed moving image after registration for an ADNI-MCI subject.}
%     \label{dicefig}
% \end{figure}

\section{Discussion}

To help avoid local minima at the finest resolution, most traditional non-linear registration algorithms follow a coarse-to-fine resolution optimization and some recent deep-learning models also adopt such pyramidal strategies at training~\cite{mok2020}. However they still rely on a prior, usually intensity-based, affine alignment to recover the largest displacement. Although it would probably allow a user to skip the first coarse levels of the pyramid, our method should not be  seen as a rival to the coarse-to-fine approach but as a more robust, anatomically grounded and less constrained alternative to the affine pre-alignment. 

In~\cite{commowick2008}, an anatomically grounded non-linear registration scheme was proposed where optimal affine transformations between homologous delineated regions were sought by iteratively optimizing an image similarity criterion. An overall transformation is then constructed by attributing the estimated affine displacement to an eroded version of the associated regions and using a log-Euclidean interpolation for a smooth transition to enforce a diffeomorphic result. Our approach, by contrast, does not require slow iterative optimization for the matching and is intended to be taken as an alternative to affine initialization rather than a non-linear registration endpoint.

Assuming that the segmentation process was successful, quality control of the registration step can easily be done just by computing an overlap measure between the transformed moving and reference segmentations.

We used segmentations based on the DKT atlas that SynthSeg and Fastsurfer pre-trained models are designed to output. This may, however, be sub-optimal as some regions are quite large (e.g. single label for superior frontal) leading to a low density of feature points in some areas. The method would likely benefit from more fine-grained and equi-distributed segmentations such as gyri delineation.

\section{Conclusion}
We presented a method to obtain a better starting point for non-linear registration than the usual affine pre-alignment. Taking advantage of the trailblazing performances of freely available pre-trained deep-learning model for fine-grained segmentation, the proposed initialization is anatomically grounded. Furthermore, thanks to the log-Euclidean polyaffine framework, the resulting transformation is diffeomorphic.  
We showed that this polyaffine initialization leads to better structural overlap, especially in the cortex, for both traditional and deep-learning non-linear registration techniques. Our experiments revealed that deep-learning registration is more sensitive to initialization and the proposed approach provides a highly efficient and effective strategy to tackle this issue. We also verified that the overall transformation is indeed proper. Finally, the proposed polyaffine estimation was more robust and faster than using an affine registration algorithm like FLIRT.


\end{document}